\documentclass[letterpaper, 10 pt, conference]{ieeeconf}  
\usepackage{graphicx}
\usepackage{bm}
\usepackage[]{algorithm2e}

\IEEEoverridecommandlockouts                              

\title{\LARGE \bf
A hierarchical behavior prediction framework at signalized intersections
}

\author{Zhen Yang$^{1}$, Rusheng Zhang$^{1}$, and Henry X. Liu$^{1}$ 
\thanks{$^{1}$Zhen Yang, Rusheng Zhang and Henry X. Liu are with Department of Civil and Environmental Engineering,
        University of Michigan, 2350 Hayward Street, Ann Arbor, MI 48109 USA.
        {email: \tt\small zhenyang, rushengz,   henryliu@umich.edu}}%
}

\begin{document}

\maketitle
\thispagestyle{empty}
\pagestyle{empty}

\begin{abstract}

Road user behavior prediction is one of the most critical components in trajectory planning for autonomous driving, especially in urban scenarios involving traffic signals. In this paper, a hierarchical framework is proposed to predict vehicle behaviors at a signalized intersection, using the traffic signal information of the intersection. The framework is composed of two phases: a discrete intention prediction phase and a continuous trajectory prediction phase. In the discrete intention prediction phase, a Bayesian network is adopted to predict the vehicle's high-level intention, after that, maximum entropy inverse reinforcement learning is utilized to learn the human driving model offline; during the online trajectory prediction phase, a driver characteristic is designed and updated to capture the different driving preferences between human drivers.  We applied the proposed framework to one of the most challenging scenarios in autonomous driving: the yellow light running scenario. Numerical experiment results are presented in the later part of the paper which show the viability of the method. The accuracy of the Bayesian network for discrete intention prediction is 91.1\%, and the prediction results are getting more and more accurate as the yellow time elapses. The average Euclidean distance error in continuous trajectory prediction is only 0.85 m in the yellow light running scenario.

\end{abstract}

\section{INTRODUCTION}

Trajectory planning plays an important role in autonomous driving, and road user behavior prediction is one of the most critical components in trajectory planning. As a key factor in the safety performance of a connected and automated vehicle (CAV), behavior prediction attracts much attention from both academia and industry. Usually, road users' behavior prediction consists of two components, discrete intention prediction and continuous trajectory prediction. Discrete intention prediction is a task of predicting the discrete maneuvers (e.g. lane change/lane keeping) that the vehicle intends to do, and continuous trajectory prediction focuses on providing a series of future locations of the target vehicle \cite{mozaffari2020deep}. Inaccurate behavior prediction of the CAV may lead to the collision to other vehicles or vulnerable road users, especially in complex urban driving scenarios.

Signalized intersections are very common in driving scenarios. When a target vehicle is approaching a signalized intersection, the behavior of the target vehicle is a complex decision incorporating many factors such as road geometry and surrounding road users, and most importantly, traffic signals. Hence, incorporating the Signal Phase and Timing (SPaT) information into the vehicle behavior prediction framework becomes a challenging yet critical topic. Although some existing studies (\cite{chen2019bayesian}\cite{ding2019online}) have made efforts in such direction, the results are preliminary. Therefore, utilizing SPaT information for intersection vehicle behavior analysis is a topic that is worthy of further exploration.


In terms of discrete intention prediction, the Bayesian network (BN) is a widely adopted methodology in existing works. Schulz et al. apply a dynamic Bayesian network to predict whether a vehicle will go straight, turn left or turn right at the intersection \cite{schulz2018interaction}. Schreier et al. build a comprehensive Bayesian network to perform discrete intention prediction and identify irrational driving behaviors in urban scenarios \cite{schreier2016integrated}. However, traffic signal information is neglected in those works. Chen et al. use BN to predict red-light-running behaviors, utilizing the trajectory information during the yellow light \cite{chen2019bayesian}. Since the intention prediction is conducted once in the paper, after the yellow light elapses 3 seconds, such discrete intention prediction results cannot be utilized in the continuous trajectory prediction during the yellow light.

For continuous trajectory prediction, inverse reinforcement learning (IRL) attracts much attention since it is explainable and interpretable. Ziebart et al. propose the idea of maximum entropy inverse reinforcement learning \cite{ziebart2008maximum}, and Levine and Koltun are among the first to apply the IRL to predict human driving behaviors \cite{levine2012continuous}. Sun et al. propose a hierarchical IRL formalism for probabilistic trajectory prediction in lane changing scenarios \cite{sun2018probabilistic}. Schwarting et al. define Social Value Orientation (SVO) when applying IRL to quantify human driving social preferences and to capture the difference between human drivers \cite{schwarting2019social}, which is validated in a highway merging scenario. However, few research works on adopting IRL at signalized intersections have been found in the literature, and this paper aims to fill in this research gap.

In this work, a hierarchical behavior prediction framework incorporating traffic signal information is proposed to predict vehicle behaviors at signalized intersections. The yellow light running scenario is adopted as an example use case, which best represents the relationship between vehicle behavior prediction and traffic signal information in urban scenarios that our model can be best applied to. This framework can be applied at the roadside infrastructure with SPaT information available, where roadside perception sensors (such as cameras and lidars) are installed that observe all vehicles approaching the intersection. It also applies to CAV which detects the front vehicle and receive SPAT message from the smart infrastructure. A BN is adopted for the discrete intention prediction to predict whether the vehicle has the intention to pass or stop at the intersection during yellow phases and the associated probability distribution. According to the results of the discrete intention prediction, continuous trajectory prediction is conducted with maximum entropy inverse reinforcement learning to produce a precisely predicted trajectory. An average human driving model is learned by IRL offline, and during the online prediction, a driver characteristic is applied to update the model to capture the individual difference between drivers (e.g. aggressive/mild).

Contributions of this work are two-fold:

\textbf{Proposing a novel hierarchical behavior prediction framework with consideration of traffic signal information:} This work presents a vehicle behavior prediction framework that combines both discrete intention prediction and continuous trajectory prediction at signalized intersections. Traffic signal information is incorporated in the prediction.

\textbf{Adopting a mixture of offline learning and online prediction strategies:} Traditional IRL can only learn an average human driving model from a static data set, which cannot represent the different driving preferences among drivers when making trajectory prediction. To solve this issue, we combine the offline learning with IRL and online updating of driver characteristics to better predict the behaviors of different drivers. 

The rest of the paper is arranged as follows: we first illustrate the yellow light running scenario discussed in this paper in the problem statement (Section II), followed by the methodology of discrete intention prediction (Section III) and continuous trajectory prediction (Section IV). Then we will show numerical experiment results (Section V), and finally Section VI concludes the paper and lays out further research directions.

\section{PROBLEM STATEMENT}

   \begin{figure}[thpb]
      \centering
      \includegraphics[scale = 0.3]{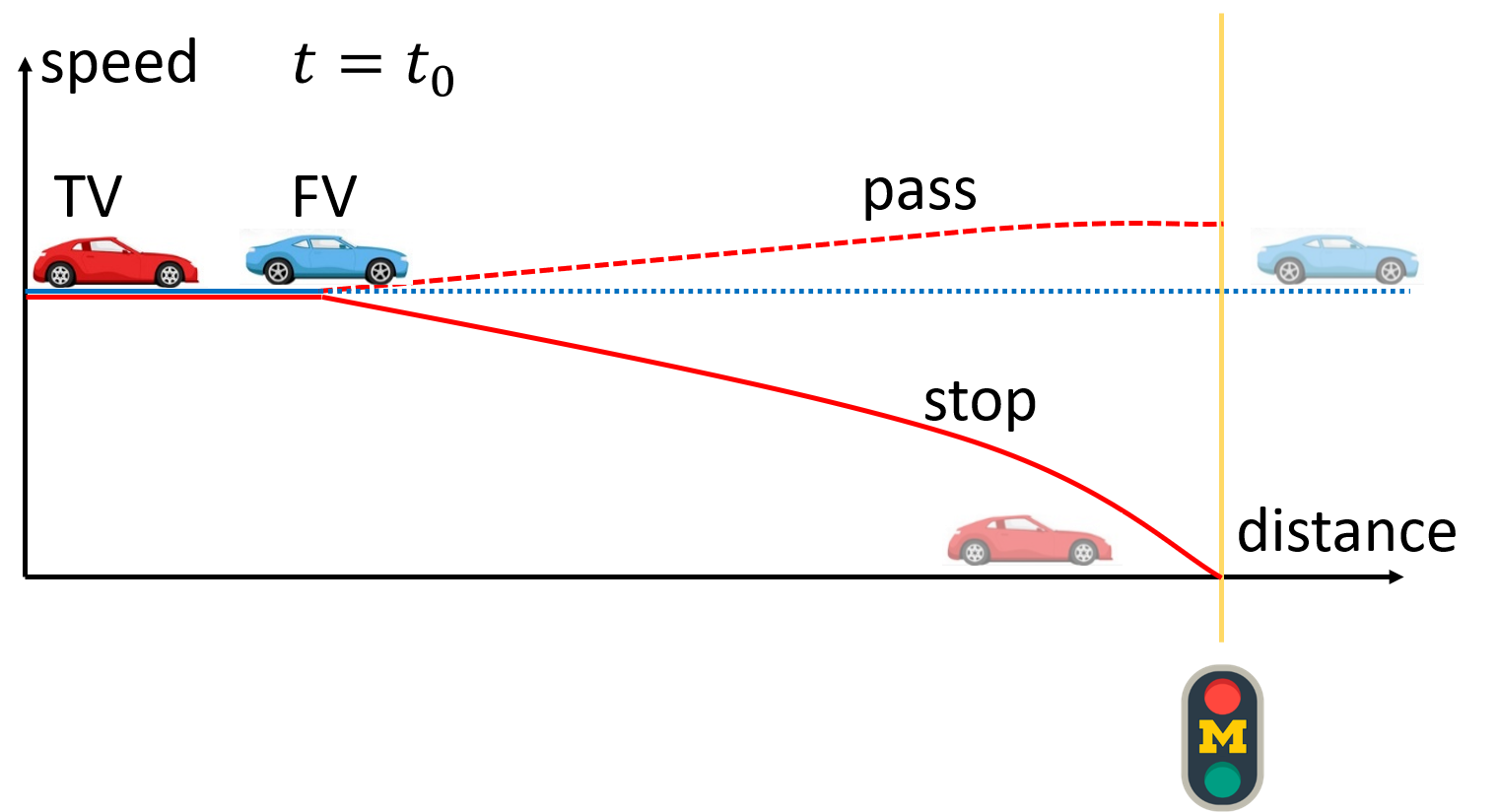}
      \caption{Yellow light running scenario}
      \label{fig:yellow}
   \end{figure}

The yellow light running scenario is taken as an example to elaborate the trajectory prediction framework in this paper, shown in Fig. \ref{fig:yellow}. The red vehicle denotes the target vehicle whose behavior needs to be predicted. Assume at time $t_0$, the behavior prediction of the target vehicle starts, and the current signal status is yellow. The target vehicle follows its front vehicle (FV), and its decision is influenced by the FV's decision. Therefore, to make the prediction of the target vehicle, the trajectory prediction of FV needs to be conducted first. In this example, the prediction result of the FV is that it cruises through the intersection (blue dash line). In terms of the behavior prediction of the target vehicle, the discrete intention prediction is conducted first to predict whether the vehicle chooses to pass or stop. According to the results of the discrete intention prediction, the continuous trajectory prediction is conducted to predict the detailed trajectory of the target vehicle (red curve).

There are three key factors in the behavior prediction of the target vehicle: the state of the target vehicle itself, the behavior of the FV, and the signal information. The state of the target vehicle includes its position, speed, acceleration, the driver's driving preference, etc. The behavior of the FV mainly refers to its speed and location since they affect the time and space headway of the car-following behavior of the target vehicle. The signal information includes the current signal status and the corresponding elapsed time. More details of the key factors and their roles in the behavior prediction are introduced in section III and section IV.

\section{DISCRETE INTENTION PREDICTION}

The discrete intention prediction module aims at providing high-level decision predictions. E.g., in the scenario of yellow light running, the goal of this module is to predict whether the target vehicle will pass or stop when facing the yellow light at each time step. As a classification problem, it not only classifies whether the target vehicle will stop or pass, but also focuses on when the target vehicle has the intention to stop or pass. In other words, the discrete intention prediction is conducted for each trajectory point during yellow phases.

\begin{figure}[thpb]
      \centering
      \includegraphics[scale = 0.7]{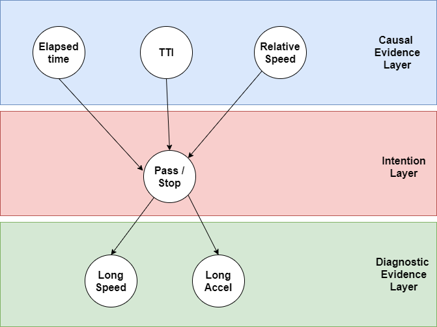}
      \caption{Bayesian network for yellow light running intention prediction}
      \label{fig:bn}
   \end{figure}

Bayesian network, as a classic classification methodology, can represent the dependencies between different random variables. Fig. \ref{fig:bn} shows the Bayesian network adopted in this paper, in which each node represents a random variable and each link denotes the dependency between two random variables. This Bayesian network includes three layers, causal evidence layer, intention layer, and diagnostic evidence layer. The middle layer, intention layer, includes the random variable of whether the vehicle chooses to pass or stop at the intersection, whose probability distribution will be predicted with this network. Causal evidence and diagnostic evidence are both observations. Causal evidence is the cause of the intention, usually observed from the environment, which includes elapsed yellow time, time to intersection (TTI), and relative speed to the front vehicle. Diagnostic evidence is the outcome of the intention, usually observed from the target vehicle, including longitudinal speed and longitudinal acceleration.

The goal of discrete intention prediction is to calculate the probability of intention ($INT$) given all causal evidence ($CE$) and diagnostic evidence ($DE$) as observations $P(INT|CE_1, ..., CE_n, DE_1, ..., DE_m)$. Here, $CE_1, ..., CE_n$ refer to the elapsed yellow time, time to intersection, and relative speed. $DE_1, ..., DE_m$ refer to the longitudinal speed and longitudinal acceleration. With the Bayesian network, the joint distribution of all the random variables (i.e. $P(INT, CE_1, ..., CE_n, DE_1, ..., DE_m)$) can be represented by some conditional distributions, shown in equation (1). The learning process of the Bayesian network is to calibrate the conditional distribution of $P(INT| CE_1, ..., CE_n)$ and $P(DE_j|INT)$ from the dataset.
$$
\begin{array}{cl}
&P(INT|CE_1, ..., CE_n, DE_1, ..., DE_m)\\
&=\frac{P(INT, CE_1, ..., CE_n, DE_1, ..., DE_m)}{\sum_{INT}P(INT, CE_1, ..., CE_n, DE_1, ..., DE_m)}\\
&=\frac{P(INT| CE_1, ..., CE_n)\Pi_{j=1}^m P(DE_j|INT)}{\sum_{INT}P(INT| CE_1, ..., CE_n)\Pi_{j=1}^m P(DE_j|INT)}\\
\end{array} \eqno{(1)}
$$

\section{CONTINUOUS TRAJECTORY PREDICTION}
In the continuous trajectory prediction module, a generic vehicle trajectory optimization problem is formulated to model the trajectory planning process, as shown in equation (2). $\bm{s}$ is the optimization variable, which represents the trajectory as a vector of trajectory points $s_i$. Each trajectory point $s_i$ at time step $i$ can be represented by ($x_i,y_i,v_i,a_i,\psi_i$), in which $x_i$ and $y_i$ are longitudinal and lateral coordinates, respectively, and $\psi_i$ is the heading angle of the vehicle, between the longitudinal axis of the vehicle and the longitudinal direction of the road. $v_i$ denotes the speed of the vehicle, and $a_i$ denotes the acceleration. $\bm{u}$ represents the initial condition and environment states, which serve as the input parameters for the optimization problem. The constraints represent vehicle dynamics, and $\bm{f}(\bm{s},\bm{u})$ is a mapping function that maps the trajectory to a feature vector, which can be different under different maneuvers and will be introduced later. The goal of the IRL is to learn the weight vector $\bm{\theta}$ associated with the feature vector $\bm{f}(\bm{s},\bm{u})$.

$$
\begin{array}{cl}
&min_{\bm{s}} \bm{\theta}^T \bm{f}(\bm{s},\bm{u})\\
&s.t. x(t+1) = x(t) + v\tau\cos{\psi}\\
&y(t+1) = y(t) + v\tau\sin{\psi}\\
&v(t+1) = v(t) + a(t)\tau\\
\end{array} \eqno{(2)}
$$

\subsection{Feature vectors}

In this part, the feature vectors that reflect the property of a trajectory with $N$ data points are introduced for the yellow light running scenario. 

(a) Speed limit: It measures the difference between the speed at each time-step $v_i$ and the speed limit $v^{lim}$.
$$
f_1 = \frac{1}{N}\sum_i (v_i - v^{lim})^2 \eqno{(3)}
$$

(b) Acceleration: It is the mean of the square of the acceleration at each time step.
$$
f_2 = \frac{1}{N}\sum_i a_i^2 \eqno{(4)}
$$

(c) Car following: $h_i$ is the headway to the leading vehicle at time step $i$. If the speed is larger than a threshold $v_{min}$, $h_i$ is the time headway. Otherwise, when the vehicle almost stops, $h_i$ is the space headway.
$$
\begin{array}{cl}
&f_3 = \frac{1}{N}\sum_i \frac{1}{h_i^2}\\
&h_i=\{\begin{array}{lr}
    \frac{d_i}{v_i} \qquad  v_i > v_{min}\\
    d_i \qquad else\\
    \end{array}\\
\end{array} \eqno{(5)}
$$

(d) Heading: This feature is the mean of the square of the heading at each time step.
$$
f_4 = \frac{1}{N}\sum_i \psi_i^2 \eqno{(6)}
$$

(e) Lateral acceleration: this feature captures the lateral acceleration at each time step.
$$
f_5 = \frac{1}{N}\sum_i (a_i \sin(\psi_i))^2 \eqno{(7)}
$$

(f) Stop position: If a vehicle chooses to stop, before the time step when the vehicle starts to launch (i.e. $i_{launch}$), the vehicle will try to approach the end of the queue $x_{queue}$. In this work, it is assumed that the time when the vehicle starts to launch and the end of the queue can be predicted accurately using shock wave theory \cite{wu2011shockwave}. The methodology of predicting these two parameters can be found in \cite{yang2019eco}.
$$
f_6 = \frac{1}{i_{launch}}\sum_i^{i_{launch}} (x_i-x_{queue})^2 \eqno{(8)}
$$

\subsection{Maximum entropy inverse reinforcement learning}
Maximum entropy inverse reinforcement learning \cite{ziebart2008maximum} is adopted to learn the human driving model offline. Consider the vehicle trajectory planning as a Markov Decision Process (MDP) with discounted cost as $\sum_{i=0}^{N-1}\gamma^i r(s_i)$, in which $\gamma$ is the discounted factor and $r$ is the reward function, the goal of IRL is to learn the reward function. If the discounted factor is taken as 1, then the discounted return is $\bm{\theta}^T \bm{f}(\bm{s},\bm{u})$, for each trajectory $\bm{s}$, so the goal is to find the weight vector $\bm{\theta}$ that maximizes the log-likelihood function $L(\bm{\theta})$, shown in equation (9). $D$ is the demonstrated trajectory data set, including m trajectories. $P(\bm{s}_j|\bm{\theta}, \bm{u}_j)$ is the probability of trajectory $s_j$ given parameter $\bm{\theta}$ and the initial condition as well as the environment state of trajectory $\bm{s}_j$ (i.e., $\bm{u}_j$), so when maximizing $L(\bm{\theta})$, the likelihood of using parameter $\bm{\theta}$ to generate all trajectories in the dataset is maximized.

$$
L(\bm{\theta}) = \frac{1}{m}\sum_{\bm{s}_j \in D} \ln P(\bm{s}_j|\bm{\theta}, \bm{u}_j) \eqno{(9)}
$$

Assuming $C_j$ is the set of trajectories generated by the Markov Chain $M^\pi$, in which $\pi$ is the maximum entropy policy. As is shown in equation (10), the maximum entropy policy finds the probability $P(\bm{s}_j|\bm{\theta}, \bm{u}_j)$ by maximizing the entropy of the probability distribution of set $C_j$, subjecting to the constraint that the expected feature vector (i.e. $\sum_{\bm{s}_k\in C_j} P(\bm{s}_k|\bm{\theta}, \bm{u}_j)\bm{f}(\bm{s}_k,\bm{u}_j)$) is equal to the empirical feature vector $\bm{f}(\bm{s}_j,\bm{u}_j)$ collected in the dataset \cite{jaynes1957information}. Besides, the summation of the probability $P(\bm{s}_k|\bm{\theta}, \bm{u}_j)$ should be equal to 1.

$$
\begin{array}{cl}
&max \sum_{\bm{s}_k\in C_j} P(\bm{s}_k|\bm{\theta}, \bm{u}_j)\log \frac{1}{P(\bm{s}_k|\bm{\theta}, \bm{u}_j)}\\
&s.t. \sum_{\bm{s}_k\in C_j} P(\bm{s}_k|\bm{\theta}, \bm{u}_j)\bm{f}(\bm{s}_k,\bm{u}_j) = \bm{f}(\bm{s}_j,\bm{u}_j)\\
&\sum_{\bm{s}_k\in C_j} P(\bm{s}_k|\bm{\theta}, \bm{u}_j)=1\\
\end{array} \eqno{(10)}
$$

After solving the problem in Equation (10), for a given trajectory $\bm{s}_j$, $P(\bm{s}_j|\bm{\theta}, \bm{u}_j)$ can be written as Equation (11).Then, the gradient of $L(\bm{\theta})$ can be calculated as Equation (12), in which $\tilde{f} = \frac{1}{m}\sum_{\bm{s}_j}\bm{f}(\bm{s}_j,\bm{u}_j)$  denotes the empirical feature vector. Thus, the gradient of $L(\bm{\theta})$ is the difference between the expected feature vector with respect to parameter $\bm{\theta}$ and the empirical feature vector calculated from the dataset. Furthermore, the expected feature vector can be approximated by the feature vector of the most likely trajectory \cite{kuderer2015learning}, shown in equation (13).

$$
P(\bm{s}_j|\bm{\theta}, \bm{u}_j) = \frac{e^{-\bm{\theta}^T \bm{f}(\bm{s}_j,\bm{u}_j)}}{\sum_{\bm{s}_k\in C_j}e^{-\bm{\theta}^T \bm{f}(\bm{s}_k,\bm{u}_j)}} \eqno{(11)}
$$

$$
\begin{array}{cl}
\nabla_{\bm{\theta}}L(\bm{\theta})&=\frac{1}{m}\sum_{\bm{s}_j\in D}\sum_{\bm{s}_k\in C_j} P(\bm{s}_k|\bm{\theta}, \bm{u}_j)\bm{f}(\bm{s}_k,\bm{u}_j)\\
&-\frac{1}{m}\sum_{\bm{s}_j}\bm{f}(\bm{s}_j,\bm{u}_j)\\
&=\frac{1}{m}\sum_{\bm{s}_j\in D}E_{P(\bm{s}_j|\bm{\theta}, \bm{u}_j)}[\bm{f}(\bm{s}_j,\bm{u}_j)]-\tilde{f}\\
\end{array} \eqno{(12)}
$$

$$
\begin{array}{cl}
E_{P(\bm{s}_j|\bm{\theta}, \bm{u}_j)}[\bm{f}(\bm{s}_j,\bm{u}_j)] &\approx \bm{f}(argmax P(\bm{s}_j|\bm{\theta}, \bm{u}_j)) \\
&= \bm{f}(argmin \bm{\theta}^T \bm{f}(\bm{s}_j,\bm{u}_j)) \\
\end{array} \eqno{(13)}
$$

Notice that since $\bm{\theta}$ represents the weight vector associated with the feature vector, all of its entries should be non-negative. However, normal gradient descent method does not guarantee that. To fix it, we make $J(\bm{\eta}) = L(e^{\bm{\eta}})=L(\bm{\theta})$, and perform gradient descent on $\bm{\eta}$. In this way, $\bm{\eta}$ can take any real number as its entries and $\bm{\theta}$ always have positive entries. According to the chain rule, the gradient of $\bm{\eta}$ is shown in equation (14).

$$
\nabla_{\bm{\eta}}J(\bm{\eta}) = \bm{\theta} \nabla_{\bm{\theta}}L(\bm{\theta})\eqno{(14)}
$$

The pseudo-code of the maximum entropy inverse reinforcement learning algorithm can be presented as follows, given a dataset $D$ of trajectories $\{\bm{s}_1,…,\bm{s}_m\}$:

\begin{algorithm}
    Compute the empirical feature vector over all trajectories $\tilde{f_0} = \frac{1}{m}\sum_{\bm{s}_j \in D}\bm{f}(\bm{s}_j,\bm{u}_j)$. Normalize the feature vector so that all the features have the same magnitude. The normalized feature vector is denoted as $\tilde{f}$.\\
    Initialize the weight vector $\bm{\theta}$. $\bm{\eta} = \log \bm{\theta}$ \\
    \While{$\frac{1}{m}\sum_{j=1}f(\bm{s}_j^{\bm{\theta}}, \bm{u}_j) - \tilde{f} > threshold$}
    {
        \For {each trajectory $\bm{s}_j$ in the dataset}
        {
            fix the environment state $\bm{u}_j$ and solve the trajectory optimization problem in (2). The optimized trajectories are denoted as $\{\bm{s}_1^{\bm{\theta}}, ..., \bm{s}_m^{\bm{\theta}} \}$ 
        }
        Gradient: $\nabla_{\bm{\theta}}L(\bm{\theta}) = \frac{1}{m}\sum_{j}f(\bm{s}_j^{\bm{\theta}}, \bm{u}_j) - \tilde{f}$ \\
        Update $\bm{\eta}$: $\bm{\eta}(k+1) = \bm{\eta}(k) + \alpha \nabla_{\bm{\theta}}L(\bm{\theta})\bm{\theta}(k)$\\
        Update $\bm{\theta}$: $\bm{\theta}(k+1) = e^{\bm{\eta}(k+1)}$
    }
 \caption{maximum entropy IRL}
\end{algorithm}

\subsection{Online trajectory prediction}
The yellow light scenario is divided into two sub-scenarios: pass or stop. Feature (a)-(e) are selected for pass sub-scenario, and feature (b)-(f) are selected for stop sub-scenario. The weight vectors $\bm{\theta}$ associated with two sub-scenarios (i.e. $\bm{\theta}_{pass}$ and $\bm{\theta}_{stop}$) are trained separately. After receiving the discrete intention prediction results of pass or stop, the corresponding features and weight vector will be utilized to make continuous trajectory prediction by solving optimization problem (2). 

During online trajectory prediction, a driver characteristic parameter $\lambda$ is defined to update the driving model learned offline. This driver characteristic parameter is utilized to capture the difference between human drivers. For example, some drivers prefer a smooth driving experience, while others may pursue driving efficiency more. To quantify such difference, features related to efficiency (i.e. speed limit feature and stop position feature) are multiplied by $\lambda$, and a feature that is related to smoothness (i.e. acceleration feature) is multiplied by $1 - \lambda$. The driver characteristic parameter is sampled during the online trajectory prediction for each driver. For each sampled $\lambda_p$, optimization problem (2) is solved to get a sampled trajectory prediction $\bm{s}_p$ associated with $\lambda_p$. Such $\bm{s}_p$ is compared with the observation of actual trajectory, and $\lambda$ is updated as $\lambda_p^*$, which is associated with the most similar trajectory $\bm{s}_p^*$ to the observed trajectory. The similarity of two trajectories with $N$ data points is measured by Euclidean Distance (ED), shown in equation (15). In this way, the weight vector $\bm{\theta}$ is adjusted by $\lambda$ for different drivers.

$$
ED = \frac{1}{N}\sum_i\sqrt{(x_i^{traj1} - x_i^{traj2})^2 + (y_i^{traj1} - y_i^{traj2})^2}\eqno{(15)}
$$

\section{NUMERICAL EXPERIMENT}

\begin{figure}[thpb]
      \centering
      \includegraphics[scale = 0.4]{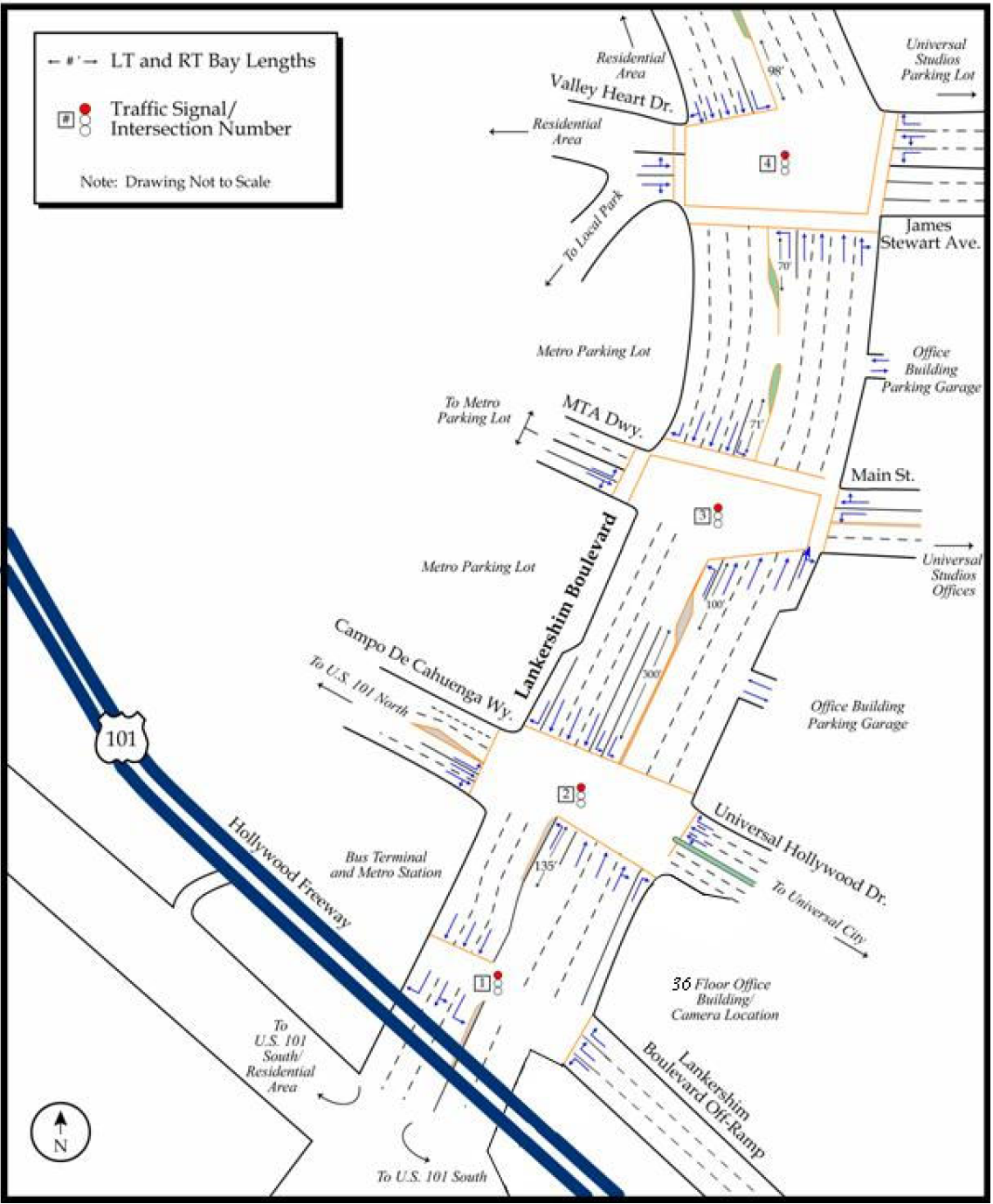}
      \caption{Road geometry of Lankershim Boulevard}
      \label{fig:ngsim}
   \end{figure}

To validate the proposed hierarchical behavior prediction framework, Next Generation Simulation (NGSIM) dataset of Lankershim Boulevard is utilized, which includes vehicle trajectory data in 10 Hz on a bidirectional urban signalized arterial, captured by cameras mounted on the roof of surrounding buildings\cite{alexiadis2004next}. The road geometry of Lankershim Boulevard is shown in Fig. \ref{fig:ngsim}. A total of 30 minutes of trajectory data are included in the full dataset, and traffic signal data associated with vehicle trajectory data are also available in the dataset.

\subsection{Experiments for discrete intention prediction}

For discrete intention prediction with the Bayesian network, 361 trajectories that experience yellow light are extracted from the NGSIM dataset. The trajectories are labeled manually. If at the end of the yellow light, the distance of the vehicle to the stop bar is larger than a threshold, all the trajectory points of this trajectory are labeled as "stop". Otherwise, the trajectory points will be labeled as "pass" for this trajectory.

To justify the performance of the Bayesian network, a naive predictor is applied for yellow light running discrete intention prediction as the baseline. The idea of the naive predictor is that if the maximum travel distance within the remaining yellow time is larger than the distance to the stop bar, the discrete intention prediction is "pass". Otherwise, the prediction result is "stop" \cite{liu2009virtual}. The naive predictor basically assumes that if the vehicle is able to pass the intersection during the yellow light, it should choose to pass.

The accuracy of the Bayesian network is 91.2\%, while the accuracy of the naive predictor is 83.2\%. It indicates that in the real world, although some vehicles can pass the intersection during the yellow light, the drivers still choose to stop for safety concerns. Notice that the discrete intention prediction is conducted trajectory-point-wise, which means that the prediction is made for each trajectory point (every 0.1 s) during yellow phases. For some cases, the discrete intention prediction is not correct at the beginning of the yellow phase, but gets corrected as time elapses.

\begin{figure}[thpb]
  \centering
  \includegraphics[scale = 0.34]{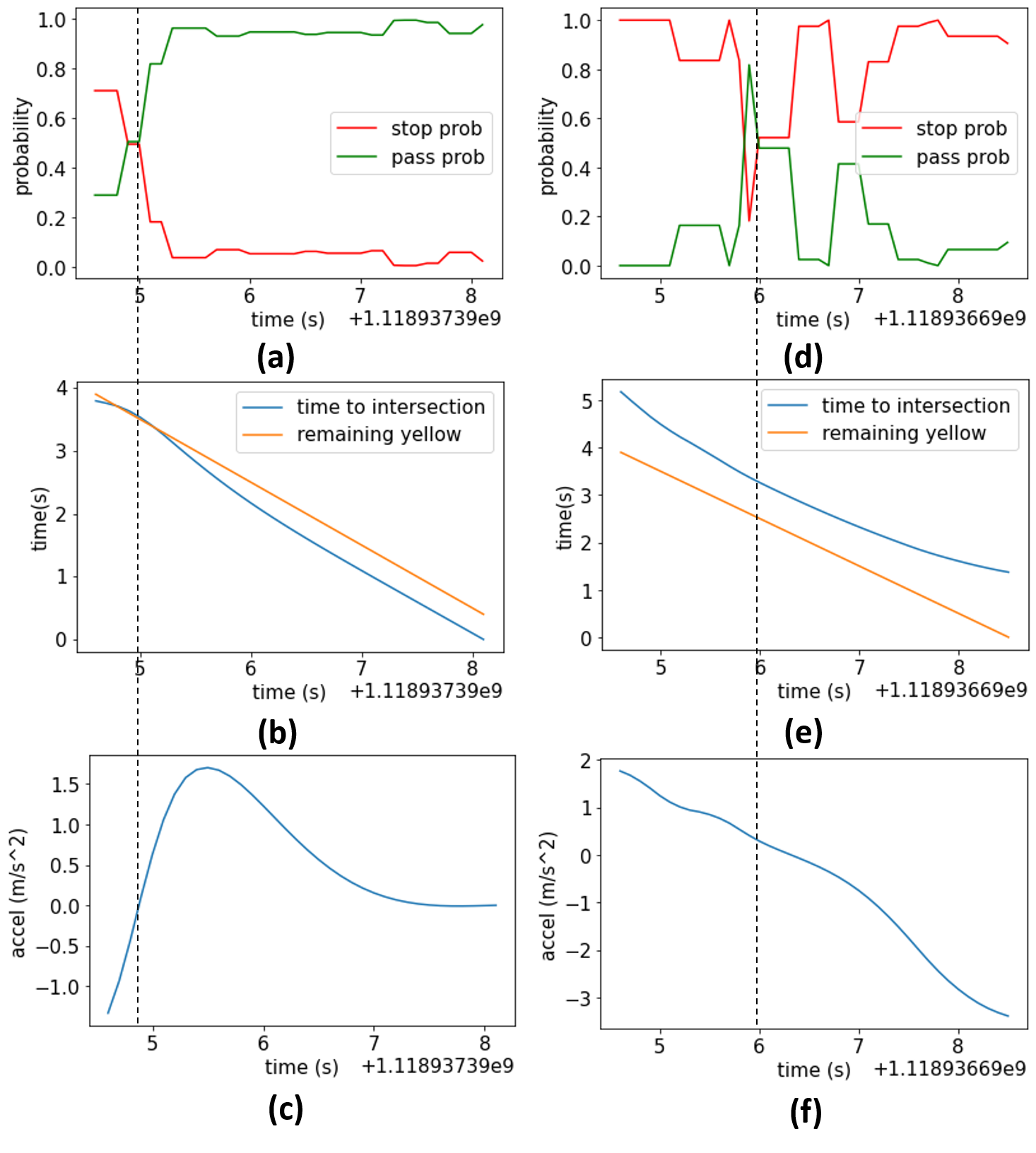}
  \caption{Case study of discrete intention prediction}
  \label{fig:intent}
\end{figure}

Fig. \ref{fig:intent} shows two specific cases that are representative (subfigure (a)-(c) and subfigure (d)-(f)) in the yellow light scenarios. The top two subfigures (i.e. (a) and (d)) are the probability-time diagrams, in which the red curves denote the probability variation of stop and the green curves denote the probability variation of pass. At each time step, the summation of the probability of stop and probability of pass should always equal to 1. To analyze the probability variation in these two specific cases, the profile of some key features as remaining time, time to intersection, and longitudinal acceleration is shown in subfigures (b), (c), (e), and (f). The middle two subfigures (i.e. (b) and (e)) show the relationship between the remaining yellow time (yellow line) and time to intersection (blue curve). When the time to intersection is larger than the remaining yellow time, the vehicle is far away from the intersection and is less likely to has the intention of pass. The bottom two subfigures (i.e. (c) and (f)) are acceleration profiles. In the first case (subfigure (a)-(c)), the probability of stop is larger than the probability of pass. The reason of that is the acceleration at the beginning of the yellow is negative, which indicates the driver may want to stop when the yellow just starts. However, since the time to intersection is quite close to the remaining yellow time, and the vehicle can actually pass the intersection. Thus, the vehicle starts to accelerate, and the prediction result changes to pass. In the second case (subfigure (d)-(f)), the prediction results fluctuate at the middle of the yellow time. Subfigure (e) shows that at the beginning, the vehicle is far away from the intersection (i.e. time to intersection is much larger than the remaining yellow time). However, the driver thinks the vehicle can pass the intersection and accelerates, so the time to intersection is getting closer and closer to the remaining yellow time. Because of that, the probability of stop goes down at the beginning. Later on, the driver realizes that it is impossible to pass the intersection without violating the red light because the distance to the intersection is still too far, then the vehicle starts to decelerate and the probability of stop rises back.

\subsection{Experiments for continuous trajectory prediction}

For continuous trajectory prediction, 60 training vehicle trajectories and 60 testing vehicle trajectories are extracted from the NGSIM dataset. In the training dataset, half of the trajectories stop at the intersection and the other half pass the intersection. The training dataset is divided into two subsets with only stop or pass trajectories, since the weight vectors of pass sub-scenario $\bm{\theta}_{pass}$ and stop sub-scenario $\bm{\theta}_{stop}$ needs to be trained separately. All testing vehicles experience yellow light when they approach the intersection.

The continuous trajectory prediction is conducted in a rolling horizon fashion. Every 0.5 seconds, the discrete intention prediction is made first, and based on the results of the discrete intention prediction, the continuous trajectory prediction is made for a prediction horizon of 3 seconds. Meanwhile, the driver characteristic $\lambda$ is also sampled and sampled prediction is generated accordingly, which are compared with the observed actual trajectory. The driver characteristic $\lambda$ is updated during the sampling process. The continuous trajectory prediction always uses the most up-to-date $\lambda$ to make predictions. It constantly makes trajectory prediction until the vehicle passes the intersection or stops at the end of the queue.

A naive predictor that assumes constant speed and constant heading in the prediction window (3 seconds)is applied as the baseline prediction. Notice that the heading here refers to the angle between the longitudinal  axis  of  the  vehicle and the longitudinal direction of the road, as is introduced before. Euclidean distance shown in equation (15) is adopted as the performance measurement. The prediction method proposed in this paper has an error of 0.85 m on average in a 3-second horizon, and the naive predictor has an error of 1.75 m.

   \begin{figure}[thpb]
      \centering
      \includegraphics[scale = 0.3]{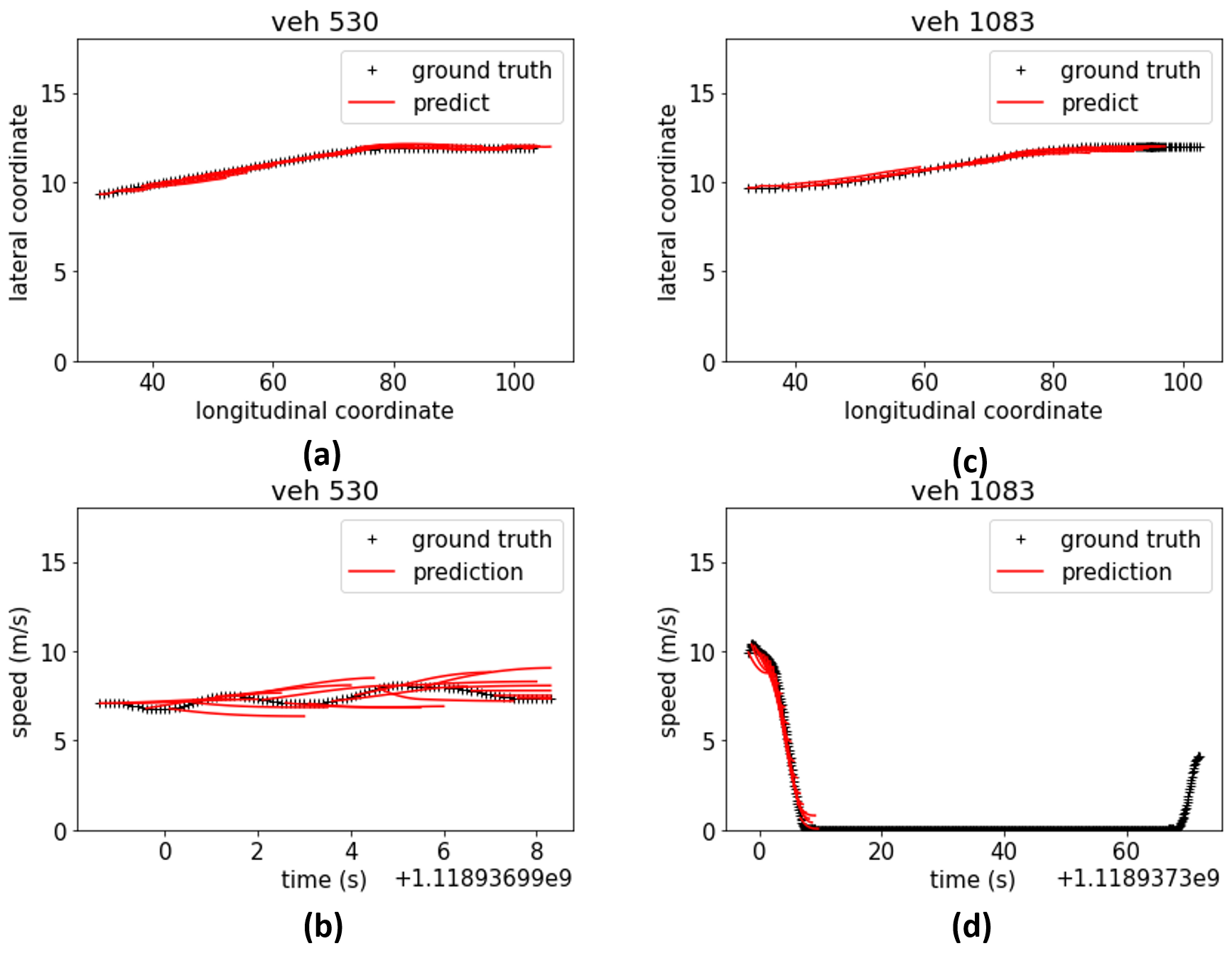}
      \caption{Case study of continuous trajectory prediction}
      \label{fig:traj}
   \end{figure}

Fig. \ref{fig:traj} shows the continuous trajectory prediction of two cases. In all four subfigures, black curves with crosses denote the actual trajectory, and the red curves denote the prediction conducted every 0.5 seconds with a 3-second prediction horizon. Subfigure (a) and (b) shows the position profile and speed profile respectively of the pass sub-scenario, and subfigure (c) and (d) show the position profile and speed profile respectively of the stop subscenario. In both sub-scenarios, the prediction is quite close to the ground truth, in terms of the position profile and speed profile.

\section{CONCLUSIONS}

In this paper, a hierarchical behavior prediction framework incorporating traffic signal information is proposed to predict vehicle behaviors at signalized intersections. The yellow light running scenario is chosen as a representative example to elaborate the framework. For discrete intention prediction, a Bayesian network is adopted, which utilizes the dependencies between causal evidence, intention, and diagnostic evidence to make predictions of whether the target vehicle chooses to stop or not. Based on the results from the discrete intention prediction, continuous trajectory prediction is then conducted accordingly. For continuous trajectory prediction, maximum entropy inverse reinforcement learning is utilized to learn average driving models for the pass and stop sub-scenarios separately. During the online prediction, the average driving models are updated by choosing the appropriate driver characteristic to capture the different driving preferences between human drivers.

In the numerical experiments, the NGSIM dataset is used to validate the hierarchical behavior prediction framework. For discrete intention prediction, the accuracy of the Bayesian network is 91.1\% and the prediction accuracy is even higher as the yellow time elapses. For continuous trajectory prediction, the prediction error in terms of Euclidean distance is 0.85 meters, in a 3-second prediction horizon. The predicted trajectories and actual trajectories are quite similar in terms of position and speed profiles, for both stop and pass sub-scenarios.

In future work, more complicated urban scenarios that involve interaction (e.g. lane change scenario) are worthy of exploration by applying the introduced framework. Besides, considering one of the most important use cases as the trajectory prediction is conducted by the smart infrastructure with bird view cameras and lidars, another important research topic could be carried out: what guidance should be sent to the CAV to help its trajectory planning, based on the trajectory prediction results. The trajectory prediction framework proposed may also benefit the criticality assessment of the CAV, to further enhance the safety performance of its trajectory planning module.

\section*{ACKNOWLEDGMENTS}
The authors would like to thank the funding support from the University of Michigan and Ford Motor Company under the Center for Smart Vehicle in a Smart World (C-SVSW) project. The views presented in this paper are those of the authors alone. 

\addtolength{\textheight}{-12cm}   






\bibliographystyle{IEEEtran}
\bibliography{IEEEabrv, ref}

\end{document}